\documentclass[journal]{IEEEtran}

\ifCLASSINFOpdf
\else
   \usepackage[dvips]{graphicx}
\fi
\usepackage{url}
\usepackage{tikz}
\usepackage{amsmath}
\usepackage{amssymb}
\usepackage{multirow}
\usepackage{amsfonts}
\usepackage{mathrsfs}
\usepackage{booktabs}
\usepackage{authblk}
\usepackage{algorithmicx}
\usepackage{algorithm}
\usepackage{booktabs}
\usepackage{algpseudocode}
\usepackage{amsthm}
\usepackage{latexsym}
\usepackage{graphicx}
\usepackage{rotating}
\usepackage{subfig}
\usepackage{color,varioref}
\usepackage{longtable}
\newtheorem{remark}{Remark}

\newtheorem{example}{Example}

\begin{document}

\title{A Momentum Accelerated Algorithm  for ReLU-based  Nonlinear Matrix Decomposition}

\author{Qingsong Wang, Chunfeng Cui, and Deren Han
\thanks{This research is supported by the National Natural Science Foundation of China (NSFC) grants  12131004, 12401415, 12471282, the R\&D project of Pazhou Lab (Huangpu) (Grant no. 2023K0603, 2023K0604), and the Fundamental Research Funds for the Central Universities (Grant No. YWF-22-T-204). (Corresponding author: Chunfeng Cui)}
\thanks{Qingsong Wang is with the School of Mathematics and Computational Science, Xiangtan University, Xiangtan, 411105, China (e-mail: nothing2wang@hotmail.com).}
\thanks{Chunfeng Cui and Deren Han  are with the LMIB, School of Mathematical Sciences, Beihang University, Beijing 100191, China (e-mail: chunfengcui@buaa.edu.cn; handr@buaa.edu.cn).}
}

\markboth{Journal of \LaTeX\ Class Files, Vol. 14, No. 8, August 2023}
{Shell \MakeLowercase{\textit{et al.}}: Bare Demo of IEEEtran.cls for IEEE Journals} 
\maketitle

\begin{abstract}
Recently, there has been a growing interest in the exploration of Nonlinear Matrix Decomposition (NMD) due to its close ties with neural networks. NMD aims to find a low-rank matrix from a sparse nonnegative matrix with a per-element nonlinear function. A typical choice is the Rectified Linear Unit (ReLU) activation function. To address over-fitting in the existing ReLU-based NMD model (ReLU-NMD), we propose a Tikhonov regularized ReLU-NMD model, referred to as ReLU-NMD-T. Subsequently,  we introduce a momentum accelerated algorithm for handling the ReLU-NMD-T model. A distinctive feature, setting our work apart from most existing studies, is the incorporation of both positive and negative momentum parameters in our algorithm.  Our numerical experiments on real-world datasets show the effectiveness of the proposed model and algorithm. Moreover, the code is available at  \url{https://github.com/nothing2wang/NMD-TM}.
\end{abstract}


\IEEEpeerreviewmaketitle

\section{Introduction}

\IEEEPARstart{M}{any} machine learning problems \cite{Bishop06} aim to identify a low-rank matrix that approximates the given matrix. Well-known methods for this include Truncated Singular Value Decomposition (TSVD) \cite{EckartY36} and Nonnegative Matrix Factorization (NMF) \cite{LeeS99, Gillis20}. Sparse nonnegative matrices are prevalent in diverse applications like edge detection \cite{BR17}, gene expression \cite{MaruyamaMMKIMA14}, social networks \cite{Hoff05}, data mining \cite{Swamy16, WangH23}, and recommender systems \cite{HanafiA21}. Nonlinear Matrix Decomposition (NMD) has gained traction due to its association with neural networks \cite{MazumdarR19, Saul22, Wang0DZHZL23}.  Given $M\in\mathbb{R}^{m\times n}$ and $r<\min(m,n)$, the essence of  ReLU-NMD involves solving the following problem:
\begin{eqnarray}
\begin{aligned}
\min_{X\in\mathbb{R}^{m\times n}}\, &\frac{1}{2}\|M-\max(0,X)\|^{2}_{F},\\
\text{s.t.}\quad\, &\text{rank}(X)=r, 
\end{aligned}\label{NMD}
\end{eqnarray}
where $\text{rank}(X)$ denotes the rank of the matrix $X$, $\|\cdot\|_{F}$ denotes the Frobenius norm of a matrix. We denote $I_{0}:=\{(i,j)\, |\, M_{ij}=0\} $ and $I_{+}:=\{ (i,j) \,|\, M_{ij}>0\}$, respectively.

We use the following sample example to illustrate the significance of this optimization problem. 
\begin{example}\label{motivation}
Given a matrix $M\in\mathbb{R}^{5\times5}$ with $M=\max(0,X)$, where 
\begin{align*}
M=\left[
\setlength{\arraycolsep}{3pt}
\begin{array}{ccccc}
3 &0 &0 &0 &0\\
0 &0 &0 &5 &4\\
0 &1 &4 &3 &0\\
0 &0 &0 &4 &5\\
5 &1 &0 &0 &0
\end{array}\right], X=\left[
\setlength{\arraycolsep}{2pt}
\begin{array}{ccccc}
3 &-1 &-4 &-3 &0\\
-5 &-1 &0 &5 &4\\
-3 &1 &4 &3 &0\\
-4 &-2 &-3 &4 &5\\
5 &1 &0 &-5 &-4
\end{array}\right]. 
\end{align*}
We know $\text{rank}(M)=5$, $\text{rank}(X)=2$ with 
\begin{align*}
X=\left[
\setlength{\arraycolsep}{2pt}
\begin{array}{ccccc}
-2 &2 &2 &1 &-2\\
-1 &-1 &1 &-2 &1
\end{array}\right]^{\mathrm{T}}\times \left[
\setlength{\arraycolsep}{2pt}
\begin{array}{ccccc}
-2 &0 &1 &2 &1\\
1 &1 &2 &-1 &-2
\end{array}\right].
\end{align*}
We transform the initially sparse nonnegative matrix $M$ into a low-rank matrix $X$ such that $M=\max(0, X)$. Subsequently, employing various low-rank matrix factorization methods becomes a viable means to alleviate the computational burden associated with this problem. 
\end{example}

The objective function outlined in \eqref{NMD} lacks both differentiability and convexity. Moreover, the inherent nonlinearity stemming from the ReLU function complicates direct problem-solving approaches. Saul \cite{Saul22} subsequently introduced an alternative version of ReLU-NMD, as follows:
\begin{eqnarray}
\begin{aligned}
\underset{X,W}{\min}\, &\  \frac{1}{2}\|W-X\|^{2}_{F},\\
\text{s.t.}\,\,\, &\   \text{rank}(X)=r,\,\,\max(0,W)=M. \label{NMD-01}
\end{aligned}
\end{eqnarray}
The algorithms proposed in \cite{Saul22, Saul23} for \eqref{NMD-01} require computing a rank-$r$ TSVD at each iteration. To circumvent this computationally intensive step, \cite{SeraghitiAVPG23} suggests replacing $W\in\mathbb{R}^{m\times n}$ with the product $UV$, where $U\in\mathbb{R}^{m\times r}$ and $V\in\mathbb{R}^{r\times n}$. Consequently, the problem can be reformulated as follows.
\begin{eqnarray}
\begin{aligned}
\underset{U,V,W}{\min}\, &\ \frac{1}{2}\|W-UV\|^{2}_{F},\\
\text{s.t.}\,\,\, &\  \max(0,W)=M. \label{NMD-MF}
\end{aligned}
\end{eqnarray}
Seraghiti et al. \cite{SeraghitiAVPG23} proposed a three-block NMD algorithm (3B-NMD)  to address this optimization problem.  The main advantage of their algorithm is that each subproblem has a closed form solution, and extrapolation techniques are also used for acceleration. Of course, other block coordinate descent (BCD)-type \cite{XuY13} algorithms and their variants, such as proximal alternating linearized minimization algorithms \cite{BolteST14, PockS16, WangH23c}    can also be applied to solve \eqref{NMD-MF}.
The experiments in \cite{SeraghitiAVPG23} showed remarkable results compared with the baselines. However,  both the $U$-  and $V$-subproblems might experience instability and the algorithm may get trapped in a worse local minima if the matrices $U$ and $V$ are allowed to grow very large before $W$ has a chance to adapt significantly as well. To counteract this, we explore a regularized version of \eqref{NMD-MF}, incorporating Tikhonov regularization \cite{GolubHO99}, represented as:
\begin{eqnarray}
\begin{aligned}
\underset{U,V,W}{\min}\, &\  \frac{1}{2}\|W-UV\|^{2}_{F}+\frac{\lambda}{2}\|U\|_{F}^{2}+\frac{\lambda}{2}\|V\|_{F}^{2},\\
\text{s.t.}\,\,\, &\ \max(0,W)=M, 
\label{NMD-T-MF}
\end{aligned}
\end{eqnarray}
where  $\lambda>0$ is a constant parameter. We denote this problem as ReLU-NMD-T. A notable advantage of this optimization problem, coupled with the Tikhonov regularization term, is both the $U$- and $V$-subproblems are strongly convex and admit unique closed-form solutions.

\textbf{Contribution.}  In this paper, we establish a robust framework in \eqref{NMD-T-MF}, and introduce a novel  algorithm (Algorithm \ref{NMD-TM}) tailored to address it. Specifically:
\begin{itemize}[]
\item \textbf{Stable model:} To combat the  unstable  challenge inherent in ReLU-NMD, we propose a more stable variant, i.e., the Tikhonov regularized version, denoted as ReLU-NMD-T.
\item \textbf{Effective algorithm:} We introduce a momentum accelerated algorithm designed for ReLU-NMD-T. This algorithm incorporates a momentum parameter, $\beta\in(0,1)$. 
Notably, contrary to the typical requirement for a positive momentum parameter for all variables,  
we set some block momentum parameters as the negative value $\beta-1$ and show promising performance.  This observation is supported by the numerical experiments in Section \ref{numerical}.
\end{itemize}
The subsequent sections of this paper are structured as follows: Section \ref{algorithm} provides an in-depth exposition of the proposed algorithm (Algorithm \ref{NMD-TM}). Section \ref{numerical} utilizes real datasets to demonstrate the stability of model \eqref{NMD-T-MF} and the efficacy of the proposed algorithm, respectively. Finally, we draw conclusions in Section \ref{conclusion}.

\section{Algorithm}\label{algorithm}
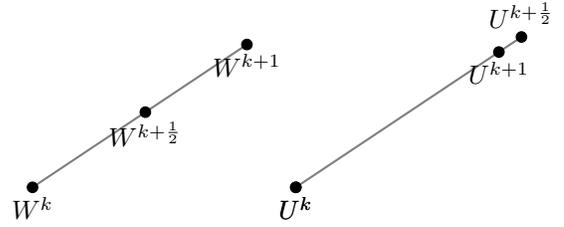
\begin{figure}
\setlength\tabcolsep{2pt}
\centering
\begin{tikzpicture}
\draw[gray, thick] (-1.5,-1) -- (0,0) -- (1.35,0.9);
\draw[gray, thick] (2,-1) -- (5,1);
\filldraw[black] (-1.5,-1) circle (2pt) node[anchor=north]{$W^{k}$};
\filldraw[black] (0,0) circle (2pt) node[anchor=north]{$W^{k+\frac{1}{2}}$};
\filldraw[black] (1.35,0.9) circle (2pt) node[anchor=north]{$W^{k+1}$};
\filldraw[black] (2,-1) circle (2pt) node[anchor=north]{$U^{k}$};
\filldraw[black] (2,-1) circle (2pt) node[anchor=north]{$U^{k}$};
\filldraw[black] (5,1) circle (2pt) node[anchor=south]{$U^{k+\frac{1}{2}}$};
\filldraw[black] (4.7,0.8) circle (2pt) node[anchor=north]{$U^{k+1}$};
\end{tikzpicture}
\caption{Illustration of the positive momentum acceleration of $W$ with $\alpha\in(0,1)$ (left) and   negative momentum acceleration of $U$   with $\beta\in(0,1)$ (right).}
\label{beta_illus}
\end{figure}

In this section, we introduce a new momentum accelerated algorithm tailored for solving problem \eqref{NMD-T-MF}. Though 3B-NMD \cite{SeraghitiAVPG23} showed commendable practical performance using $\alpha=0.7$ for variables $W$ and $X$ in solving \eqref{NMD-T-MF},  
this momentum value remains relatively conservative compared to the commonly employed parameters of $0.9$ or $0.99$ \cite{Nesterov1983, Lin2020book}. 
This motivates us to explore the potential of increasing the $\beta$ value and seek better numerical performance.
Fortunately, we can pursue this goal by setting negative momentum parameters, or equivalently, applying a convex combination step, for variables $U$ and $V$. Specifically, 
we implement the momentum acceleration step as follows: 
\begin{itemize}
    \item For  variables $W$ and $X$, we let  $\alpha\in(0,1)$ and 
    \begin{eqnarray}
       & W^{k+1}\leftarrow W^{k+\frac12}+\alpha(W^{k+\frac12}-W^{k}), \\
       &  X^{k+1}\leftarrow X^{k+\frac12}+\alpha(X^{k+\frac12}-X^{k}).
    \end{eqnarray} 
    This step is an extrapolation step.
    \item For  variables $U$ and $V$, we set  $\beta\in(0,1)$ and 
        \begin{eqnarray}
        & U^{k+1}\leftarrow  U^{k+\frac12}+(\beta-1)(U^{k+\frac12}-U^{k}), \label{U_momentum}\\
      & V^{k+1}\leftarrow V^{k+\frac12}+(\beta-1)(V^{k+\frac12}-V^{k}). \label{V_momentum}
     \end{eqnarray} 
    This step is a convex combination step. 
\end{itemize} 
{See Fig. \ref{beta_illus} for an illustration.} 
Algorithm \ref{NMD-TM} summarizes the positive and negative momentum acceleration strategies, referred to as NMD-TM. When $\lambda=0$ and $\beta=1$ in (7) and (8), the NMD-TM algorithm reduces to the 3B-NMD algorithm. 

We elucidate the rationale behind opting for the momentum parameter $\beta-1$ for variables $U$ and $V$ instead of another positive momentum parameter as follows: 
\begin{itemize}
\item Compared to the 3B-NMD \cite{SeraghitiAVPG23}, our proposed method can adopt a larger   $\alpha$ and improve the numerical performance in practice.
\item The convex combination steps in \eqref{U_momentum} and \eqref{V_momentum} enable us to pull the variable back towards the last iteration, which may decrease the possible large distance between $U^{k+\frac12}$ (resp. $V^{k+\frac12}$) and $U^{k}$ (resp. $V^{k}$)  caused by the closed form solutions of the subproblems.
\end{itemize}

\begin{algorithm}[t]
\caption{A momentum accelerated algorithm for ReLU-NMD-T (NMD-TM)}
\label{NMD-TM}
{\bfseries Input:} $M$, $r$, and   $\alpha, \beta\in(0,1)$, $\lambda>0$, $I_{+}$, $I_{0}$, and $K$. \\
{\bfseries Initialization:} $U^{0}$,  $V^{0}$, $X^{0}=U^{0}V^{0}$, and set $W_{i,j}^{k}=M_{i,j}$ for $(i,j)\in I_{+}$ and $k=0,1$.
\begin{algorithmic}[1] 
\For {$k=0,1,\dots K$} 
\State $W_{i,j}^{k+\frac12}=\min(0,X_{i,j}^{k})$ for $(i,j)
\in I_{0}$.
\State $W^{k+1}\leftarrow W^{k+\frac12}+\alpha(W^{k+\frac12}-W^{k})$.
\State $U^{k+\frac12}=\text{argmin}_{U}\frac{1}{2}\|W^{k+1}-UV^{k}\|^{2}_{F}+\frac{\lambda}{2}\|U\|_{F}^{2}$.
\State $U^{k+1}\leftarrow U^{k+\frac12}+(\beta-1)(U^{k+\frac12}-U^{k})$.
\State $V^{k+\frac12}=\text{argmin}_{V}\frac{1}{2}\|W^{k+1}-U^{k+1}V\|^{2}_{F}+\frac{\lambda}{2}\|V\|_{F}^{2}$.
\State $V^{k+1}\leftarrow V^{k+\frac12}+(\beta-1)(V^{k+\frac12}-V^{k})$.
\State $X^{k+\frac12}=U^{k+1}V^{k+1}$.
\State $X^{k+1}\leftarrow X^{k+\frac12}+\alpha(X^{k+\frac12}-X^{k})$.
\EndFor
\end{algorithmic} 
{\bfseries Output:}  $U=U^{k+1},V=V^{k+1}$.
\end{algorithm}	
\begin{remark}
Three key parameters in Algorithm 1 are the regularization weight $\lambda$ and momentum values $\alpha$ and $\beta$. 
\begin{itemize}
\item   $\lambda$  is a parameter to make the model more stable. It can be assigned any small positive value and typically does not necessitate manual adjustments during experiments.
\item Concerning the momentum parameters $\alpha$ and $\beta$, we set $\alpha=\beta$ for simplicity in numerical experiments. And we find  $\beta=0.95$ yields favourable practical performance. 
\end{itemize}
\end{remark}

\section{Numerical experiments}\label{numerical} 
We apply the proposed NMD-TM\footnote{\url{https://github.com/nothing2wang/NMD-TM}} (Algorithm \ref{NMD-TM}) to address the optimization problem \eqref{NMD-T-MF} and compare its performance with several algorithms: A-NMD (Aggressive momentum NMD) \cite{SeraghitiAVPG23}, 3B-NMD\footnote{\url{https://gitlab.com/ngillis/ReLU-NMD}} \cite{SeraghitiAVPG23}, EM-NMD (Expectation-Maximation NMD) \cite{Saul22}, and A-EM (Accelerated variant of EM-NMD) \cite{Saul23}. All experiments are conducted using MATLAB  on a PC equipped with an Intel CORE i7-14700KF @ 3.40GHz and 32GB RAM. 

Without loss of generalization, we set $\lambda=0.0001$, $U^{0}\in\mathbb{R}^{m\times r},V^{0}\in\mathbb{R}^{r\times n}$ as the best  rank-$r$ approximation of $M$ \cite{SeraghitiAVPG23}, $X^{0}=U^{0}V^{0}$,  and $W^{0}_{i,j}=\min(0,X_{i,j}^{k})$ for $(i,j)
\in I_{0}$ and zeros otherwise. We display 
\begin{eqnarray}
\text{Tol}:=\|M-\max(0,X)\|_{F}/\|M\|_{F}-\text{Tol}_{\min}, \label{real_tol}
\end{eqnarray}
where $\text{Tol}_{\min}$ is the smallest relative error obtained by all compared algorithms.

\subsection{Setting of $\beta$} \label{beta_set}

We first discuss the choice of $\beta$ by the  $28\times 28$ grayscale images from the MNIST handwritten digits dataset.

\begin{figure}
\setlength\tabcolsep{2pt}
\centering
\begin{tabular}{c}
\includegraphics[width=0.9\columnwidth]{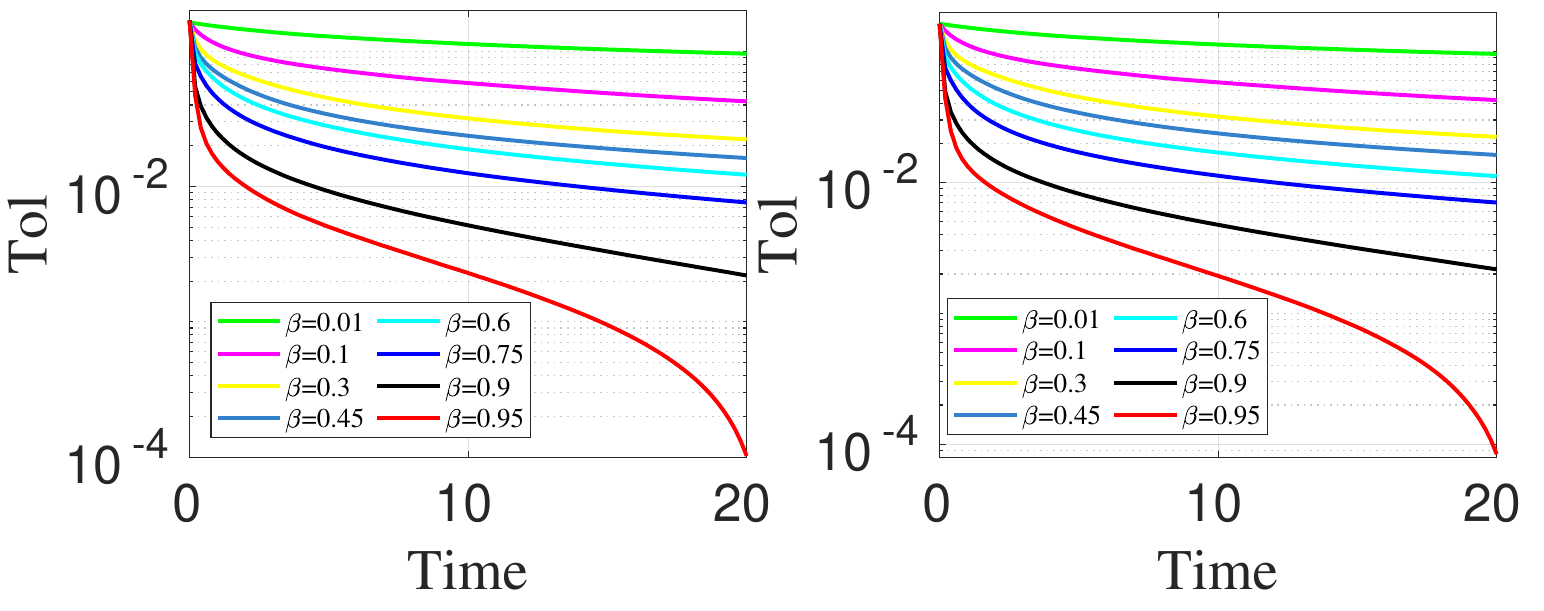}
\end{tabular}
\caption{Numerical experiments of   MNIST with $5000$ images ($500$ images of each digit) under different $\beta$ by the NMD-TM algorithm. Left: $r=30$. Right: $r=45$.}
\label{diff_beta}
\end{figure}

\begin{figure}
\setlength\tabcolsep{2pt}
\centering
\begin{tabular}{c}
\includegraphics[width=\columnwidth]{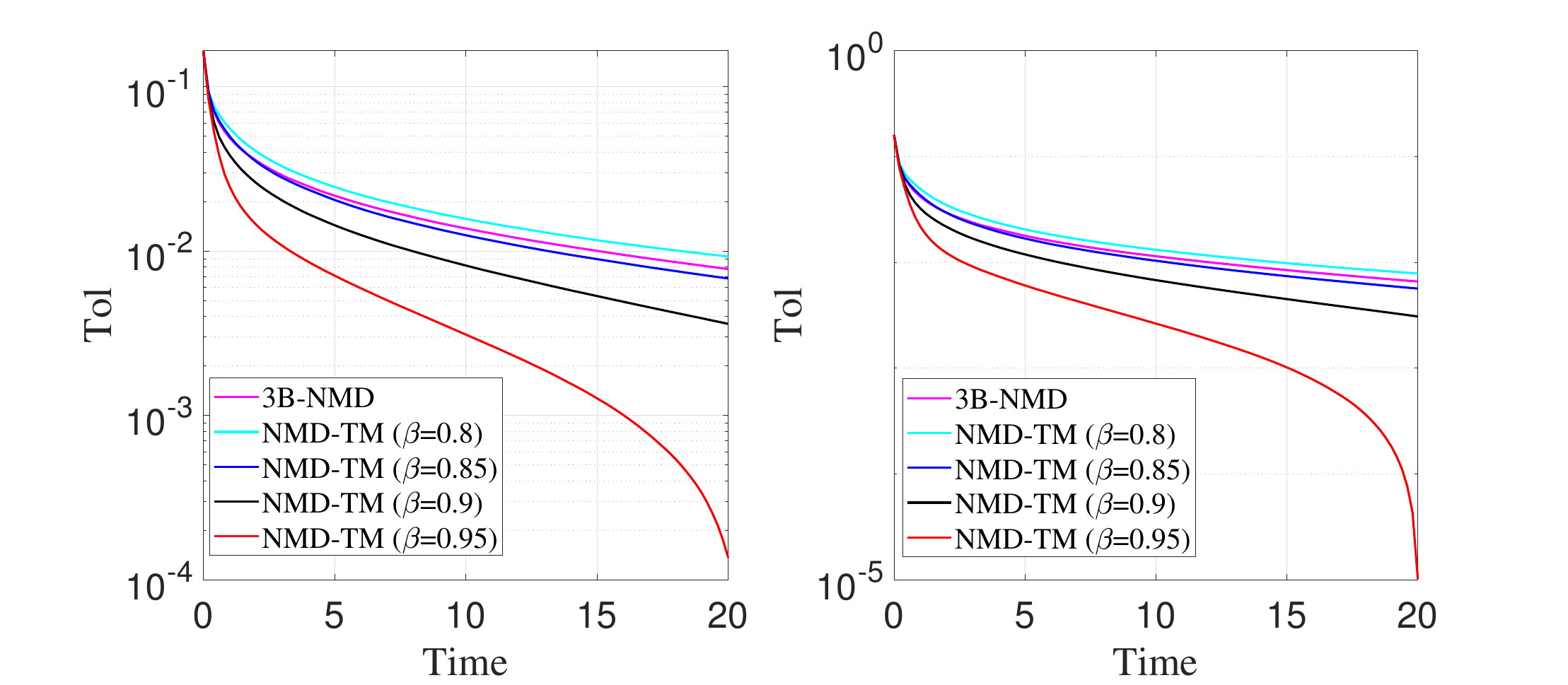}
\end{tabular}
\caption{Numerical experiments of   MNIST with $5000$ images ($500$ images of each digit). Left: $r=30$. Right: $r=50$.}
\label{beta_setting}
\end{figure}
We use  $\alpha=\beta\in\{0.01, 0.1, 0.3, 0.45, 0.6, 0.75, 0.9, 0.95\}$ and present the numerical performance. The results in Fig. \ref{diff_beta}  verify that the algorithm always converges and a larger parameter $\beta$ may produce better performance.  We also enumerate $\beta\in\{0.8, 0.85, 0.9, 0.95\}$ and $r\in\{30, 50\}$ to evaluate the numerical performance of the proposed algorithm and compare it with 3B-NMD. Fig. \ref{beta_setting} depicts that NMD-TM performs better than 3B-NMD and as $\beta$ increases, the numerical efficacy improves progressively. Notably, when $\beta=0.95$, the numerical performance reaches its peak. For ease of implementation, in subsequent numerical experiments, we will employ $\beta=0.95$.

\subsection{MNIST dataset}

\begin{figure}
\setlength\tabcolsep{2pt}
\centering
\begin{tabular}{c}
\includegraphics[width=\columnwidth]{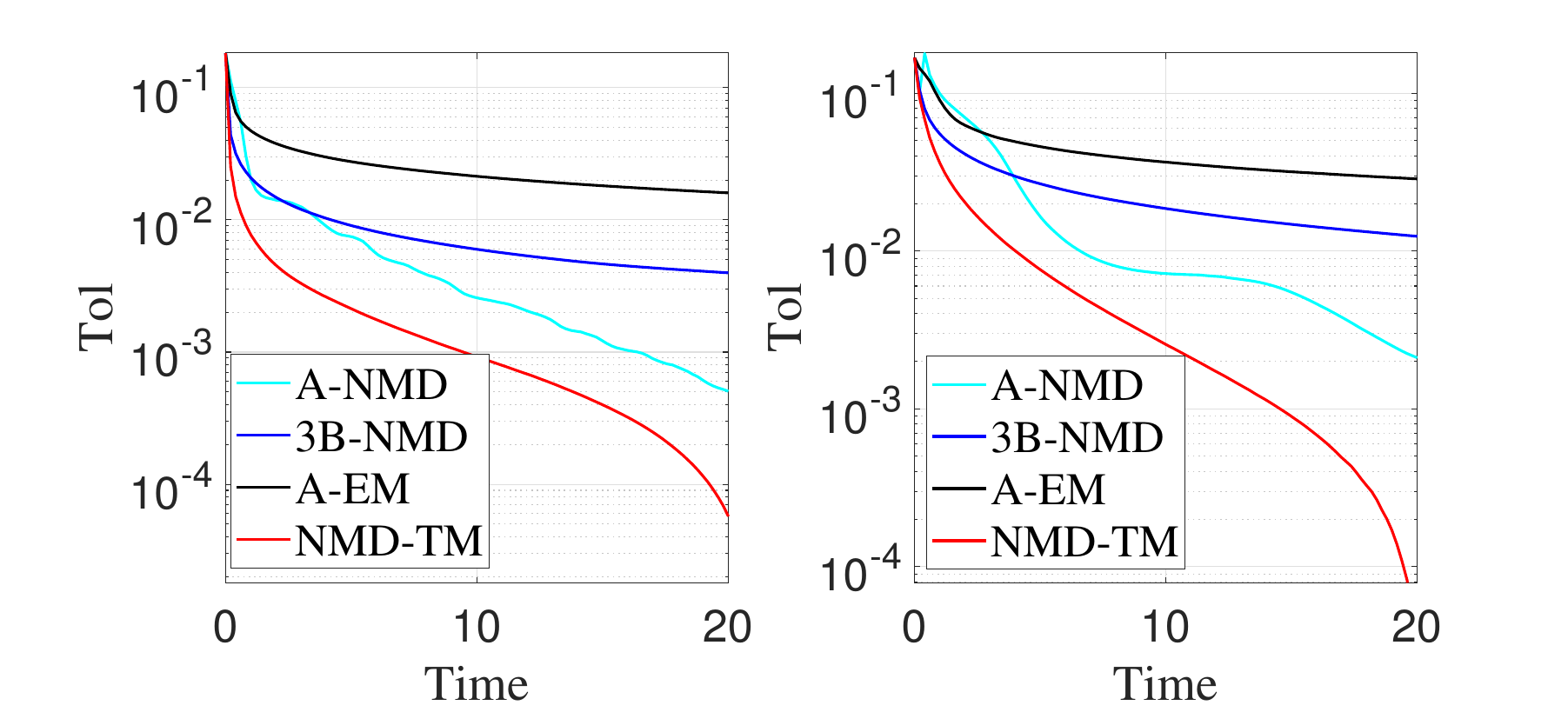}\\
(a) Left: $m=1000$. Right: $m=10000$.\\
\includegraphics[width=\columnwidth]{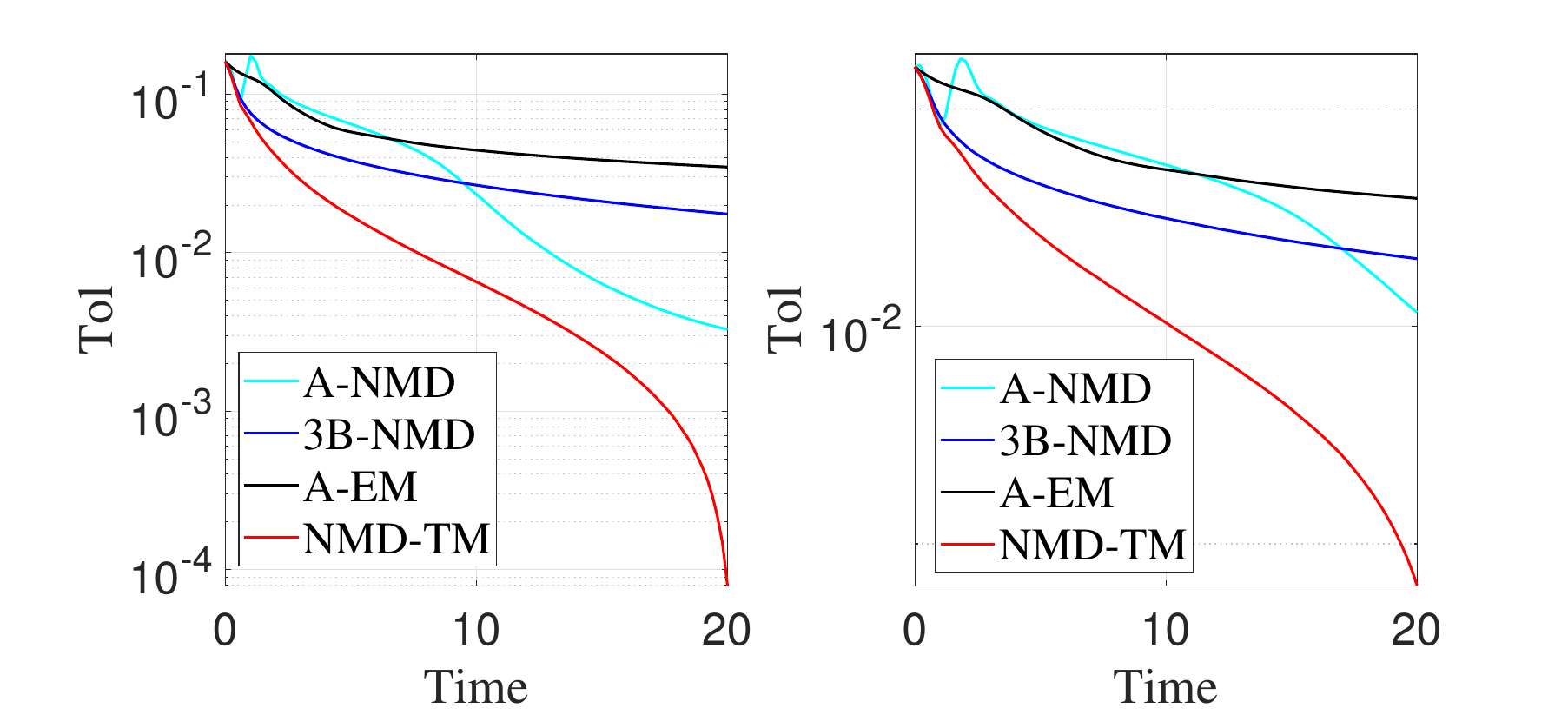}\\
(b) Left: $m=30000$. Right: $m=50000$.\\
\end{tabular}
\caption{Numerical experiments with different sample numbers of the  MNIST data set with rank $r=40$.}
\label{real_mnist_01_tol}
\end{figure}

In this section, we conduct experiments using subsets of the MNIST dataset comprising $1000$ (consisting of $100$ images per digit), $10000$, $30000$, and $50000$ images, respectively. 
Fig. \ref{real_mnist_01_tol} showcases the numerical outcomes constrained within a time limit of 20 seconds.  Right from the outset, our algorithm exhibits superior performance, and this advantage becomes increasingly pronounced as time progresses.

\begin{figure}
\setlength\tabcolsep{2pt}
\centering
\begin{tabular}{c}
\includegraphics[width=\columnwidth]{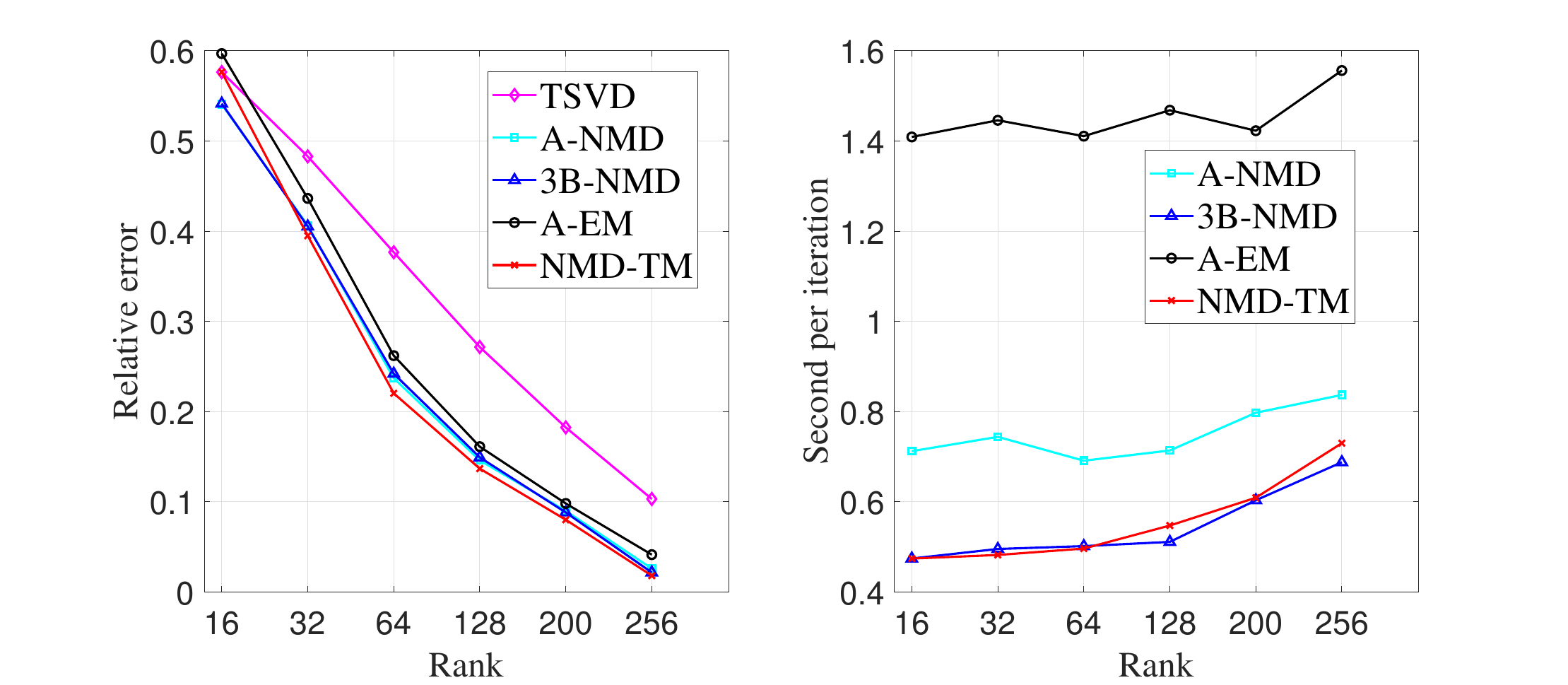}
\end{tabular}
\caption{Final relative error on $m=50000$ images from MNIST data set, after 20 seconds and average iteration time for increasing value of the rank $r$. Left: Relative error. Right: Time (seconds).}
\label{real_mnist_ranks}
\end{figure}

We continue to use $m=50000$ with increasing ranks ($r=16, 32, 64, 128, 200, 256$) to show the robustness of our algorithm. 
From Fig. \ref{real_mnist_ranks}, our proposed NMD-TM consistently yields superior results.

\subsection{Compression of sparse NMF basis}
In this subsection,  we explore another application of ReLU-NMD: the compression of sparse nonnegative dictionaries, specifically those generated by NMF \cite{LeeS99}. We use two data sets\footnote{http://www.cad.zju.edu.cn/home/dengcai/Data/FaceData.html}, i.e., the ORL data set with $m=4096, n=400$ and the YaleB data set with $m=1024,n=2414$, to illustrate this application. The NMF decomposition, denoted as $M\approx UV$ where both $U$ and $V$ are nonnegative, enables the extraction of sparse facial features represented by the columns of $U$. We perform a rank-$100$ NMF\footnote{\url{https://gitlab.com/ngillis/nmfbook/}} \cite{LeplatAG19} on the ORL dataset, resulting in $U\in\mathbb{R}^{4096\times 100}$, yielding a nonnegative sparse matrix. Similarly, for the YaleB dataset, a rank-$81$ NMF generates $U\in\mathbb{R}^{1024\times 81}$ as a nonnegative sparse matrix.

\begin{figure}
\setlength\tabcolsep{2pt}
\centering
\begin{tabular}{c}
\includegraphics[width=\columnwidth]{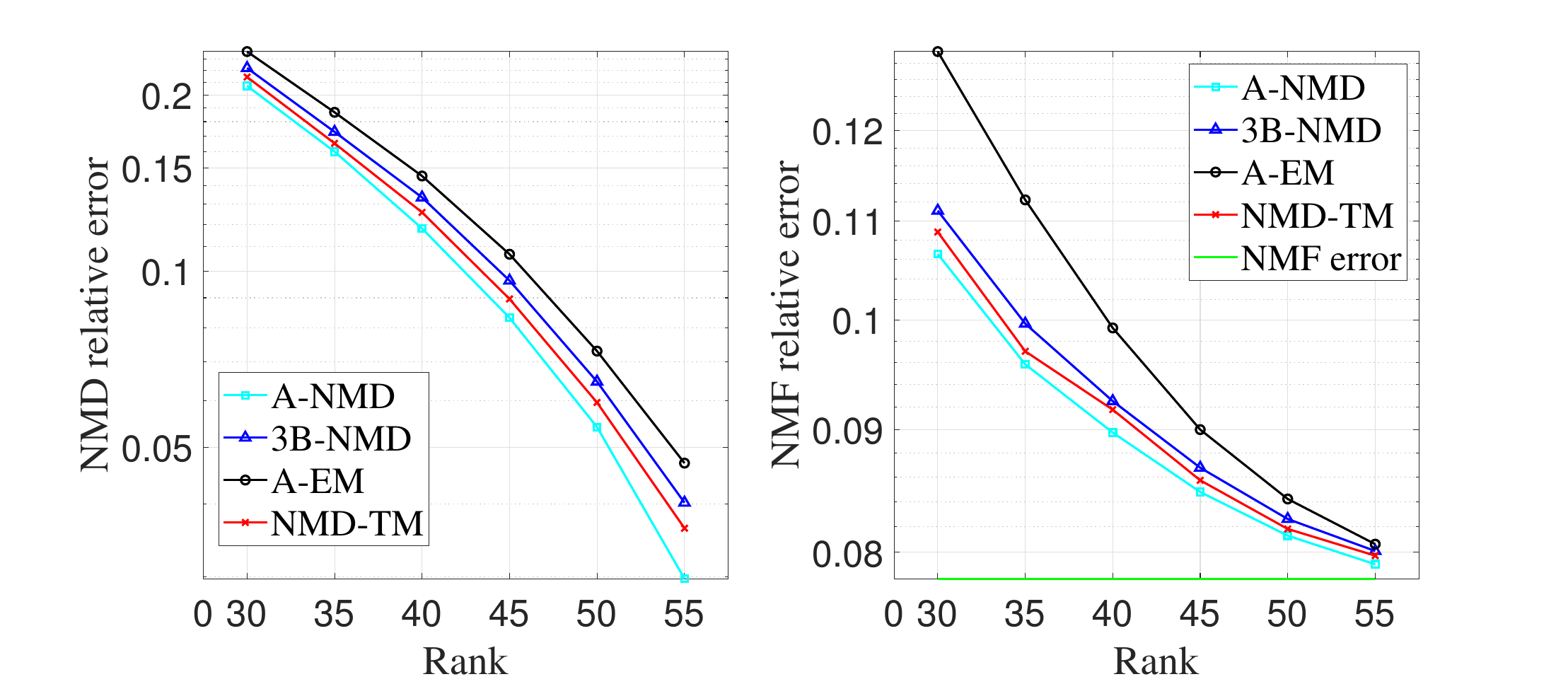}\\
(a) Compression of the ORL data set.\\
\includegraphics[width=\columnwidth]{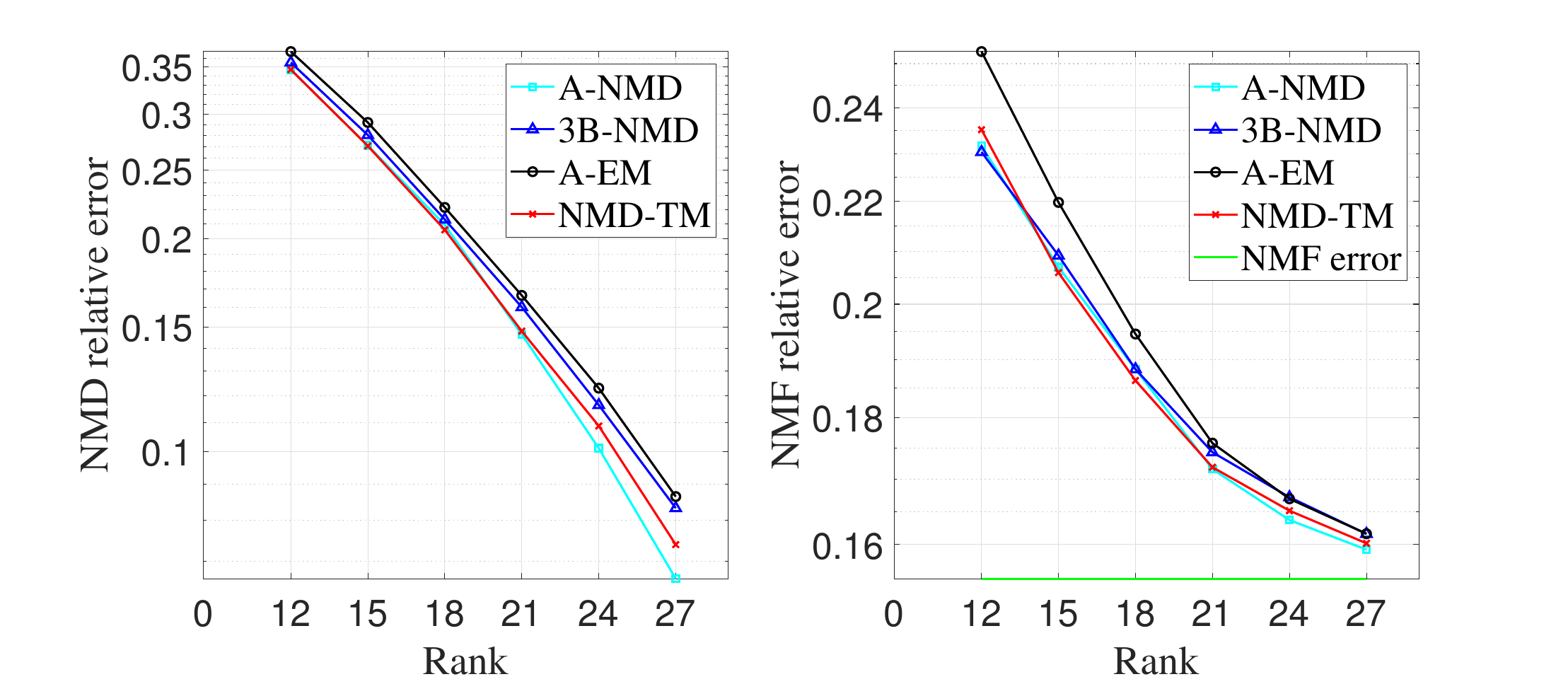}\\
(b) Compression of the YaleB data set.
\end{tabular}
\caption{Left: the error on the NMF basis $U \ge 0$. Right: the NMF error after $U$ is replaced by its approximation. See \eqref{error_nmd}.}
\label{real_yaleB_ranks}
\end{figure}

\begin{figure}
\setlength\tabcolsep{3pt}
\centering
\begin{tabular}{ccc}
\includegraphics[width=0.22\textwidth]{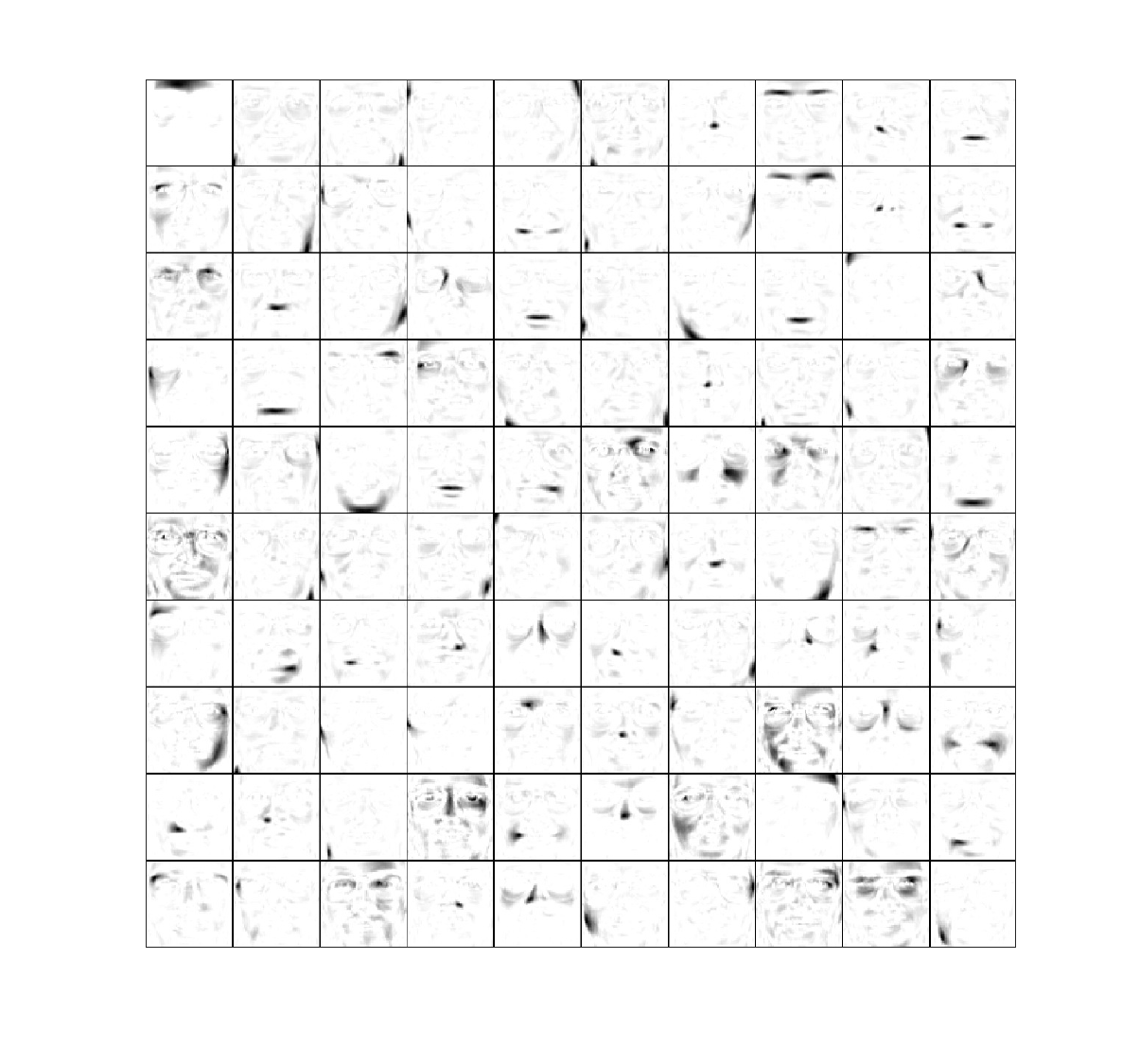}&
\includegraphics[width=0.22\textwidth]{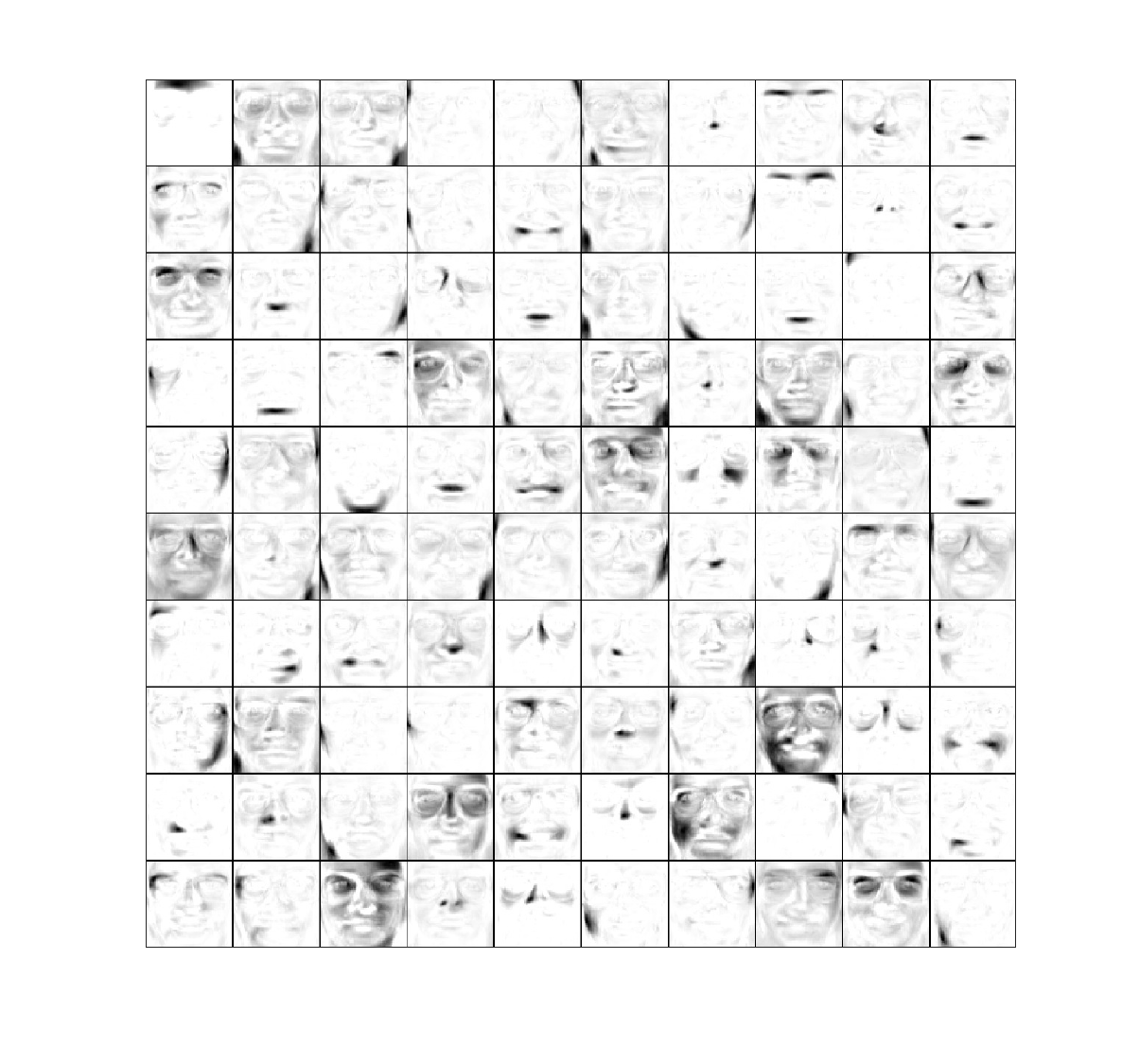}\\
(a) Original $r=100$ &(b) TSVD $r=55$ \\
\includegraphics[width=0.22\textwidth]{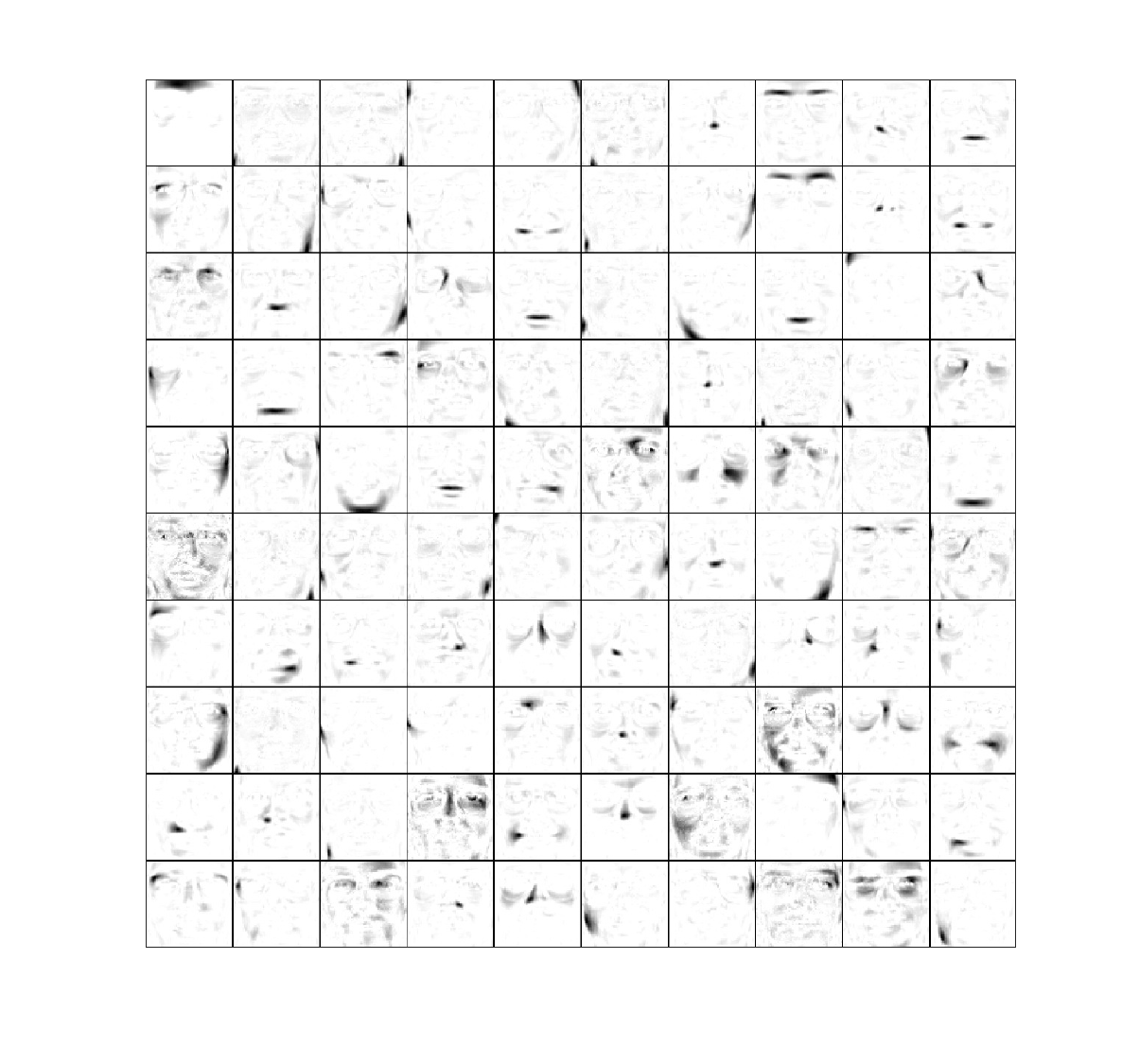}&
\includegraphics[width=0.22\textwidth]{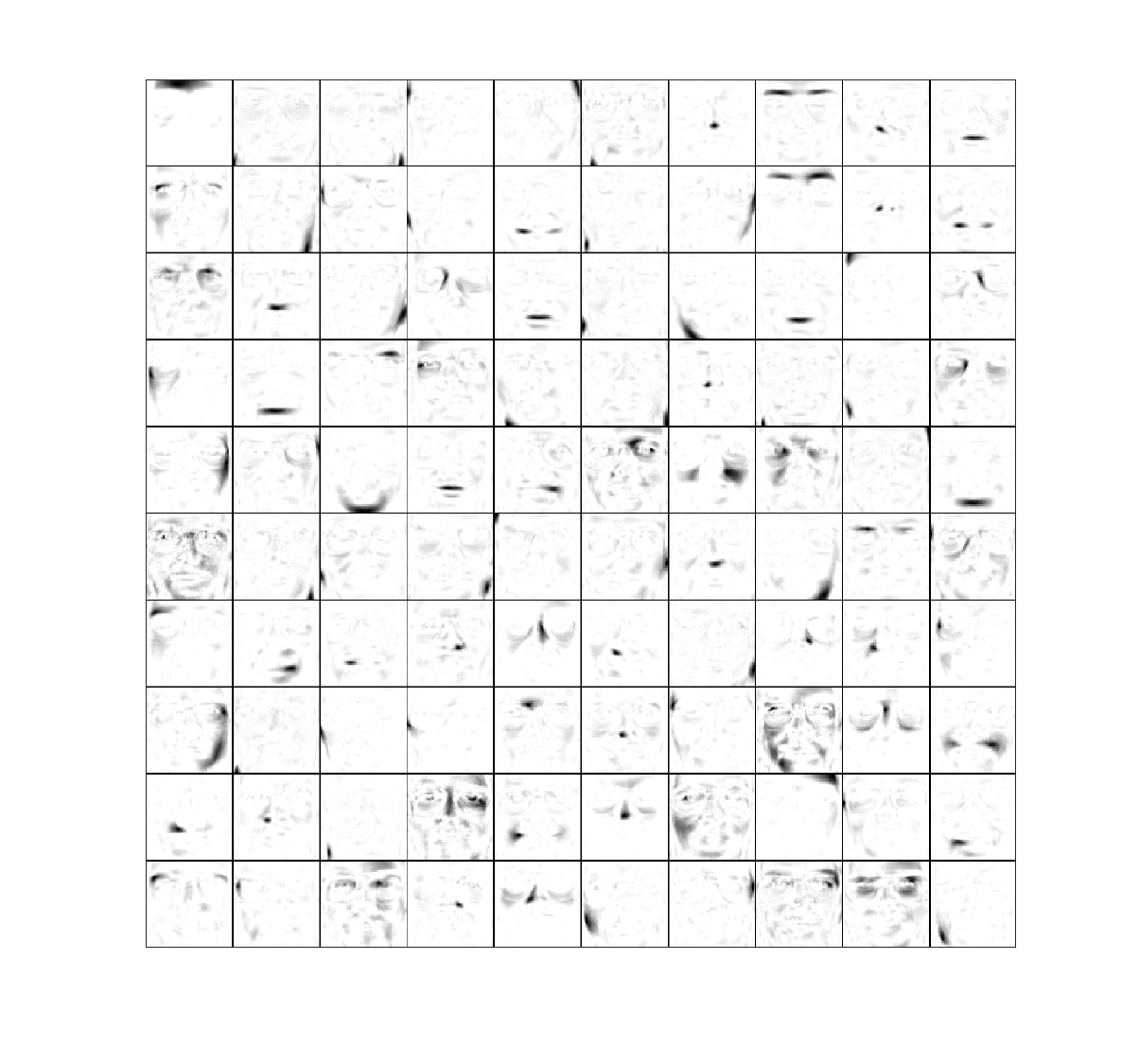}\\
(c) A-EM $r=55$ &(b) A-NMD $r=55$ \\
\includegraphics[width=0.22\textwidth]{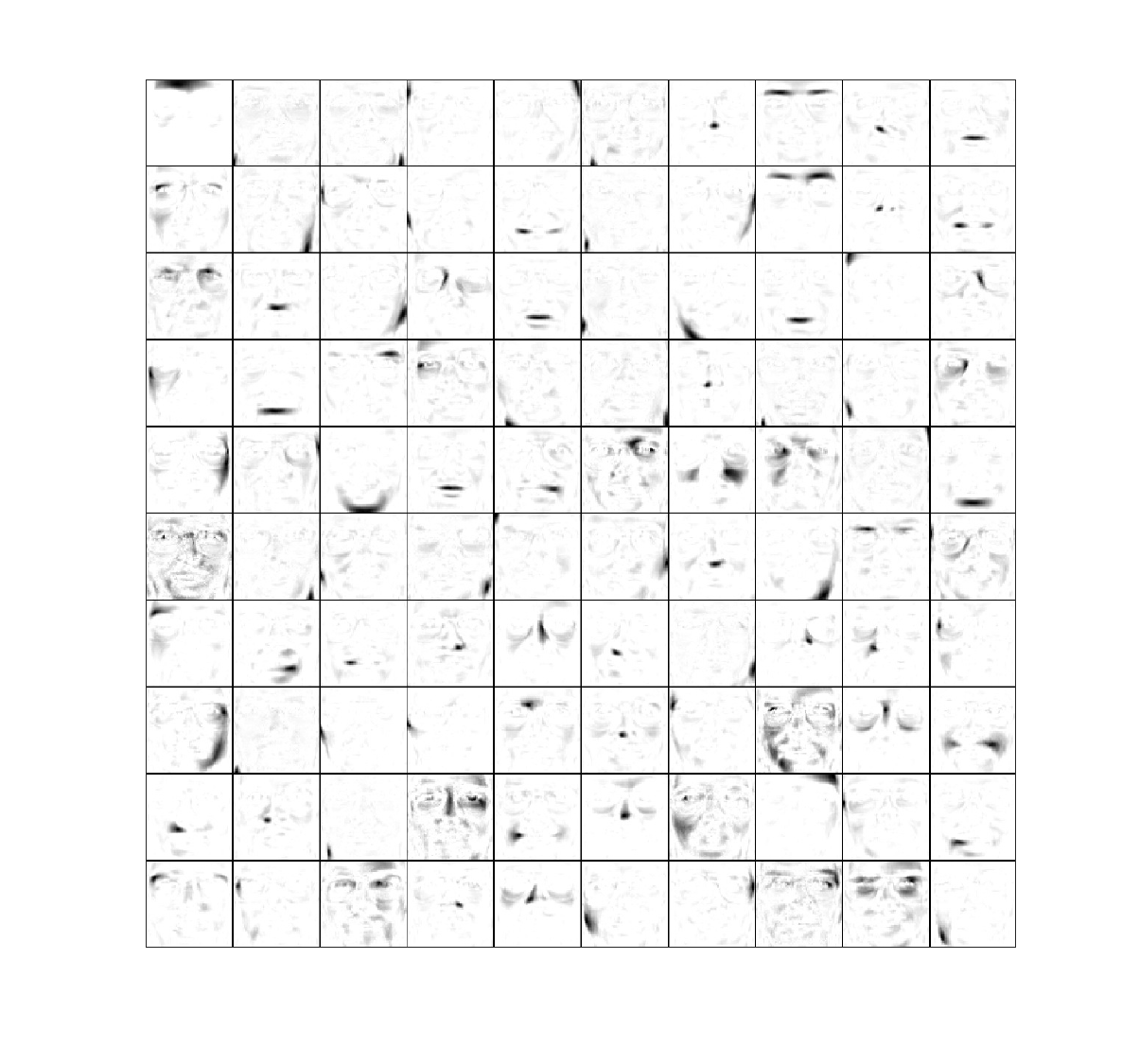}&
\includegraphics[width=0.22\textwidth]{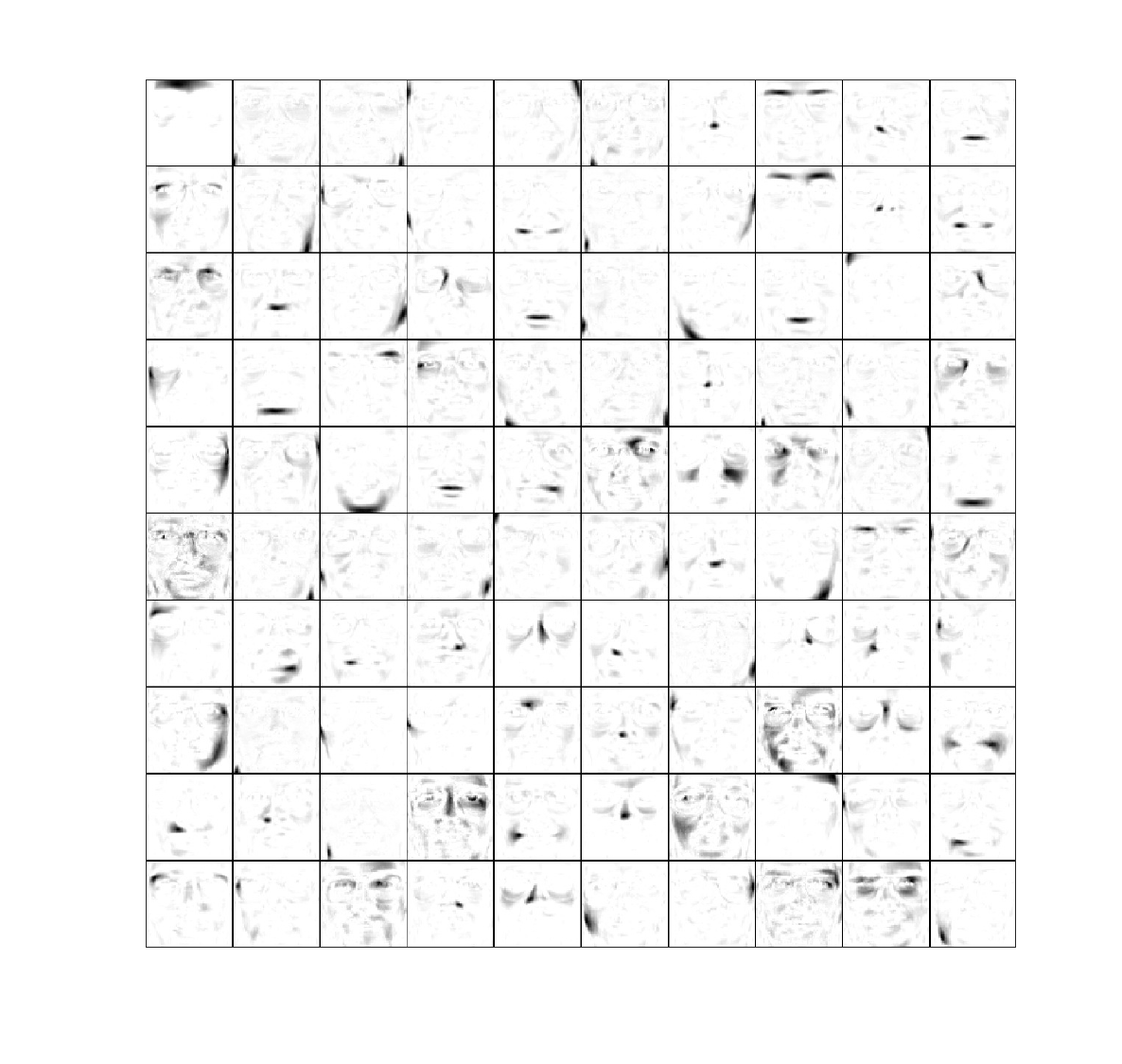}\\
(e) 3B-NMD $r=55$ &(f) NMD-TM $r=55$ \\
\end{tabular}
\caption{Original factor $U$ of NMF for ORL data set, with rank-$r=100$ and low-rank reconstruction by TSVD, A-EM, A-NMD, 3B-NMD, and NMD-TM with fixed rank $r=55$.}
\label{orl_res}
\end{figure}

Fig \ref{real_yaleB_ranks} shows the detailed numerical results for the ORL and YaleB datasets. Notably, further compressing the NMF factor $U$ using TSVD proves ineffective, as indicated in these figures. However, ReLU-NMD circumvents this limitation by effectively approximating such sparse full-rank matrices. Additionally, we assess the error of the compressed NMF by
\begin{eqnarray}
\text{Tol}_{\text{NMF}}:=\min_{\hat{V}\ge0}\frac{\|M-\max(0,\hat{U}\hat{V})\|_{F}}{\|M\|_{F}}, \label{error_nmd}
\end{eqnarray}
where $\hat{U}=\max(0,U)$. As can be seen from the right half of the two figures, as the rank increases, we can get better approximations. 
Our proposed NMD-TM demonstrates commendable performance, second only to A-NMD which needs more time in each iteration. 
Additionally, Fig. \ref{orl_res} presents an instance of a rank $r = 55$ reconstruction derived from the original rank $r = 100$ NMF factor for the ORL dataset. This example further confirms the superior accuracy of all ReLU-NMD models in approximating the dataset.

\section{Conclusion and Future Work}\label{conclusion}
In this paper, we focus on the ReLU based nonnegative matrix decomposition (NMD). We first adopted the Tikhonov regularization to improve the stability of the ReLU-NMD model. Then we proposed an accelerated algorithm with the combination of positive and negative momentum parameters tailored for solving the ReLU-NMD-T problem. Numerical experiments were presented to show the efficiency of our proposed algorithm. Several problems can be considered for future work, such as the convergence analysis and the ReLU based nonnegative tensor decomposition \cite{WangCH23, KoldaB09}.
\bibliographystyle{IEEEtran}
\bibliography{NMDTM_Ref}

\end{document}